\documentclass[10pt,twocolumn,letterpaper]{article}

\usepackage[pagenumbers]{cvpr} 

\definecolor{cvprblue}{rgb}{0.21,0.49,0.74}
\usepackage[pagebackref,breaklinks,colorlinks,allcolors=cvprblue]{hyperref}
\usepackage{graphicx}
\usepackage[linesnumbered,ruled,vlined]{algorithm2e}
\usepackage{hwemoji}
\usepackage{dsfont}
\usepackage{multirow}
\usepackage{pgf-pie}
\usepackage{microtype}
\usepackage[accsupp]{axessibility}  

\def\paperID{***} 
\def\confName{CVPR}
\def\confYear{2026}

\title{DriverGaze360: OmniDirectional Driver Attention with Object-Level Guidance}
\author{
Shreedhar Govil$^1$\and
Didier Stricker$^{1,2}$\and
Jason Rambach$^1$ \and \
$^1$German Research Center for Artificial Intelligence (DFKI) \\$^2$Rhineland-Palatinate Technical University (RPTU)\\
{\tt\small shreedhar.govil@dfki.de, didier.stricker@dfki.de, jason.rambach@dfki.de}
}

\begin{document}
\maketitle
\begin{abstract}

Predicting driver attention is a critical problem for developing explainable autonomous driving systems and understanding driver behavior in mixed human-autonomous vehicle traffic scenarios. Although significant progress has been made through large-scale driver attention datasets and deep learning architectures, existing works are constrained by narrow frontal field-of-view and limited driving diversity. Consequently, they fail to capture the full spatial context of driving environments, especially during lane changes, turns, and interactions involving peripheral objects such as pedestrians or cyclists. In this paper, we introduce DriverGaze360, a large-scale 360$^\circ$ field of view driver attention dataset, containing $\sim$1 million gaze-labeled frames collected from 19 human drivers, enabling comprehensive omnidirectional modeling of driver gaze behavior. Moreover, our panoramic attention prediction approach, DriverGaze360-Net, jointly learns attention maps and attended objects by employing an auxiliary semantic segmentation head. This improves spatial awareness and attention prediction across wide panoramic inputs. Extensive experiments demonstrate that DriverGaze360-Net achieves state-of-the-art attention prediction performance on multiple metrics on panoramic driving images. Dataset and method available at \url{https://dfki-av.github.io/drivergaze360}.

\end{abstract}    
\section{Introduction}
\label{sec:intro}



\begin{figure}[thbp]
    \centering
    \includegraphics[width=.8\linewidth]{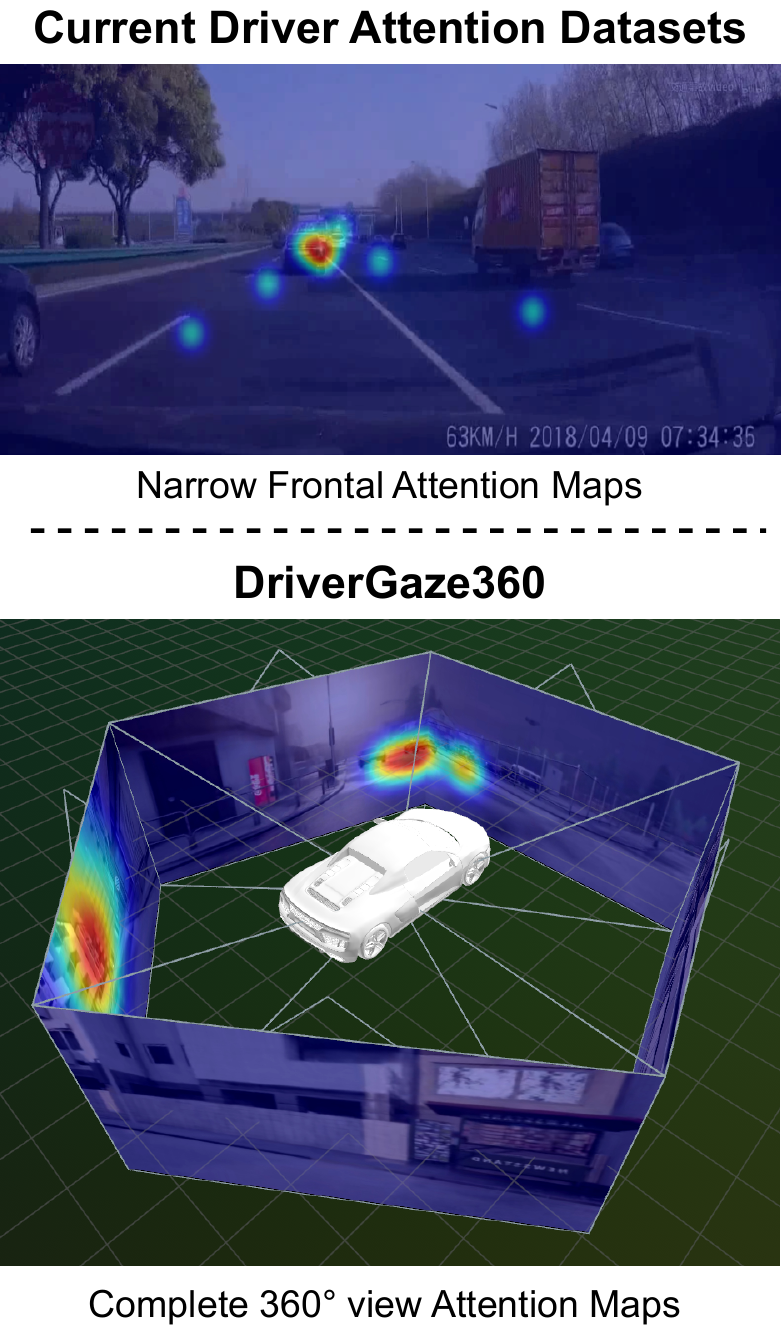}
    \caption{Existing methods predict driver attention only within a narrow frontal field of view, limiting understanding of gaze behavior. DriverGaze360 captures the full 360° field of view, enabling analysis of comprehensive gaze dynamics. In this example, the driver uses the mirrors to examine the rear-left while turning—--a behavior not modeled in prior datasets.}
    \label{fig:teaser}
\end{figure}

Safe driving requires awareness of surroundings, constant monitoring of road and traffic, and alertness to react to unexpected events~\cite{kinnear2015battle}. Advanced driving assistance systems (ADAS) such as forward collision warning and lane departure warning aim to prevent the consequences of distracted driving. The use of such technologies is promising but the overall effects are still unknown~\cite{bates2021known, cabralladas}. Hence, it becomes imperative to build ADAS systems that can predict where a driver looks at during driving and alert the driver if needed~\cite{accidentavoid, assisteddriving}.

Predicting driver attention also plays an important role for autonomous vehicle development~\cite{avdriving, perifovea, salobjseg}. Gaze behavior encodes critical cues for decision-making and reaction timing. Modeling where and what a driver is looking at enables the design of explainable-AI systems that can anticipate human intent in mixed human–AI traffic scenarios~\cite{zhou2025towards}. 

Over the past decade, several large-scale datasets~\cite{bdda, drveeye, lbw, dada2000} and learning-based models~\cite{fblnet, scout_map, medirl,dadanet,zhou2025towards} have advanced the study of driver attention. However, prior work exclusively employs narrow, forward-facing cameras that capture only a fraction of the scene. This limited field of view restricts modeling of attention during complex maneuvers such as lane changes, turns, and interactions with lateral or rear entities like pedestrians or cyclists. Moreover, most existing datasets focus on brief, safety-critical events, offering limited insight into continuous, multi-directional gaze dynamics. The absence of omnidirectional data restricts progress toward holistic driver attention understanding.

To address this gap, we introduce DriverGaze360, a large-scale 360$^\circ$ field-of-view driver attention dataset that enables comprehensive omnidirectional modeling of gaze behavior. Collected in a controlled simulation environment, DriverGaze360 provides fine-grained control over peripheral objects, lane changes, and participant behavior. As shown in~\Cref{fig:teaser}, our dataset captures realistic gaze patterns such as rear-view monitoring---patterns that are unseen in existing work.

Simultaneously, we introduce DriverGaze360-Net, a vision transformer (ViT) based attention prediction model that jointly learns attention maps and the semantic categories of attended objects (vehicles, pedestrians, cyclists, and others) through an auxiliary segmentation head. This joint formulation improves spatial awareness and robustness when predicting sparse panoramic attention distributions. Extensive experiments demonstrate that DriverGaze360-Net achieves state-of-the-art performance across multiple saliency and segmentation metrics on panoramic data and generalizes effectively to real-world driving datasets.

We summarize the contributions of this paper as follows: 
\begin{enumerate}
    \item \textbf{DriverGaze360}: The first large-scale omnidirectional driver attention dataset collected from eye tracking of human drivers in a simulated environment, spanning diverse and safety-critical scenarios including lane changes, turns, and peripheral agent interactions.
    \item \textbf{DriverGaze360-Net}: a transformer-based architecture that jointly predicts attention maps and attended objects through an auxiliary semantic segmentation head, enhancing spatial awareness and outperforming the current state-of-the-art models. 
    \item \textbf{Attended Object Extraction}: a fixation-semantic fusion pipeline that maps gaze distributions to object instances, yielding object-level attention annotations used as supervision for DriverGaze360-Net.
\end{enumerate}

\section{Related Work}
\label{sec:relatedwork}

\subsection{Driver Attention Datasets}

Over the last few years, there have been several works that build large scale datasets for capturing driver attention in various scenarios, thereby advancing understanding of human visual behavior in driving. 

The DR(eye)VE~\cite{drveeye} dataset was among the first to align captured gaze behavior with visual driving scenes, providing a foundation for learning-based attention prediction. LBW~\cite{lbw} expanded data collection to a broader driver pool under real-world conditions, while IVGaze~\cite{ivgaze} focused on in-vehicle gaze capture, but lacked corresponding driving footage. The BDD-A dataset by Xia et al.~\cite{bdda} focused on safety-critical events like emergency braking and traffic congestion. DADA-2000~\cite{dada2000} extended this to traffic accidents, including 2,000 videos and corresponding gaze data, thereby shifting the focus towards safety-critical and accident scenarios, offering large-scale annotated clips that expose the temporal dynamics of driver awareness under high-risk events.

Although these datasets have been instrumental, their scope remains limited. Most employ forward-facing cameras, representing only a fraction of the driver's visual field. Consequently, they under-represent attention behaviors associated with lateral or rear spatial contexts—such as lane changes, merging, or monitoring pedestrians in blind spots. To address these constraints, we present DriverGaze360, a human driver-synthetic scene 360$^\circ$ driver attention dataset designed to capture full-surround gaze behavior. Unlike previous frontal-view datasets, DriverGaze360 records panoramic driving environments rendered in simulation, enabling the modeling of attention beyond the windshield and full control over the captured driving scenarios.~\Cref{tab:dataset_compare} summarizes how DriverGaze360 differs from existing datasets.

\begin{table*}
\small
  \centering
  \caption{Comparison of DriverGaze360 with existing driver attention datasets.}
  \begin{tabular}{c|c|c|c|c|c}
    \toprule    
    \textbf{Dataset} & \textbf{360$^\circ$ FoV} & \textbf{\# Hours} & \textbf{Scenarios} & \textbf{\# Subjects} & \textbf{Data Collection} \\
    \midrule
    DR(eye)VE~\cite{drveeye} & ❌ & 6 & Regular Driving & 8 & Real driving \\
    LBW~\cite{lbw} & ❌ & 7 & Regular Driving & 28 & Real driving \\
    BDD-A~\cite{bdda} & ❌ & 4 & Busy Intersections, Emergency Breaking & 1,228 & Watching videos \\
    DADA-2000~\cite{dada2000} & ❌ & 6 & Driving Accidents & 20 & Watching videos \\
    \hline
    DriverGaze360 (ours) & ✅ & 9 & Regular Driving, Critical Situations & 19 & Simulated driving \\
    \bottomrule
  \end{tabular}
  \label{tab:dataset_compare}
\end{table*}

\subsection{Driver Attention Prediction}
Driver attention prediction has advanced alongside the rise of large-scale driver-attention datasets. Early work such as DR(eye)VE~\cite{drveeye} employed a multi-sensor CNN pipeline using RGB, semantic segmentation, and optical flow to approximate human gaze. BDD-A~\cite{bdda} leveraged AlexNet~\cite{alexnet} features with ConvLSTMs to capture short-term temporal cues from front-facing video. SCAFNet~\cite{dadanet} is a method of simultaneous processing and fusion of RGB and semantic images. SAGE~\cite{sage} treats all road users as a single foreground class, adding it to saliency ground truth, and performing foreground-background classification. ASIAFNet~\cite{asiaf} combines short-temporal motion features with object-level attention estimation, using bounding box detection and binary classification. While  MEDIRL~\cite{medirl} used maximum entropy depth inverse reinforcement learning. More recent approaches explore richer context: FBLNet~\cite{fblnet} introduces a feedback loop to simulate driver experience and uses dual CNN-Transformer encoders; SCOUT+~\cite{scout_map} encodes bird's‐eye‐view road geometry via a map encoder and couples it with image embeddings to improve driver attention prediction at safety‐critical scenarios and intersections. 

However, these methods operate on narrow field‐of‐view forward imagery and thus fail to represent the full 360$^\circ$ attention field that real drivers deploy. We show their performance degrades on panoramic inputs, limiting applicability to omnidirectional perception. In contrast, DriverGaze360‐Net pairs each panoramic frame with object-level semantic labels (e.g., vehicles, pedestrians, traffic signals) to identify attended entities, yielding stronger saliency performance and improved real‐world generalization. 

\section{Building the DriverGaze360 Dataset}

In this section, we introduce our setup for gaze data collection of human drivers in simulation environments. Our system includes a driving simulator, eye tracking setup, scenario player, and a post processing step to build attention maps. Leveraging this system with 19 driver subjects, we build the first 360$^\circ$ driver attention dataset. The dataset spans a diverse range of scenarios, facilitating and advancing future research on omnidirectional attention prediction.

\subsection{Data Collection Setup}

\begin{figure*}[thbp]
    \centering
    \includegraphics[width=.95\linewidth]{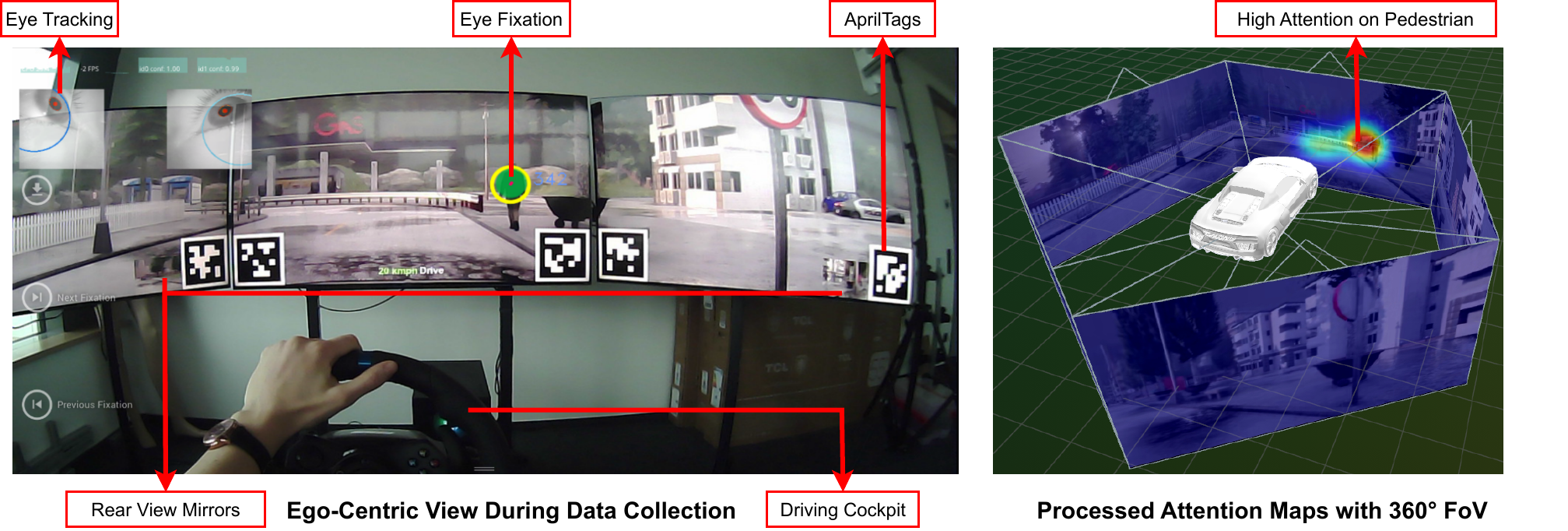}
    \caption{Experimental Setup. (Left) Egocentric perspective during a data collection with rainy weather and a pedestrian. (Right) Resulting attention maps with a 360$^\circ$ FoV.}
    \label{fig:experiment_running}
\end{figure*}

We use the CARLA simulator~\cite{dosovitskiy2017carla} due to its open-source support, reproducibility, and broad use in autonomous-driving research. CARLA provides configurable camera, LiDAR, and control interfaces, and supports flexible scenario generation. We employ both its native scenario tools and SCENIC~\cite{fremont2023scenic} to produce varied traffic, agent behaviors, and weather conditions.

To approximate a naturalistic driving field of view, the setup employs three large front displays and two picture-in-picture displays serving as rear-view mirrors with 72$^\circ$ FoV each, forming a contiguous 360$^\circ$ FoV. This configuration enables acquisition of wide-angle gaze patterns, including peripheral and backward attention shifts. The setup is shown in \Cref{fig:experiment_running} (left).

We capture the eye gaze and the driver's egocentric perspective through the Pupil Core~\cite{pupilcore} eye-tracking glasses. The device operates at 120 Hz for eye-tracking and provides a synchronized egocentric video at 30 Hz. Participants operate a cockpit system with steering, throttle, and brake, plus gear control. A heads-up display shows speed and current gear. AprilTags~\cite{olson2011tags} rendered in the simulated scene enable calibration between eye-tracker coordinates and simulator space.

\subsection{Traffic Scenarios}

To mitigate sim-to-real bias and ensure broad attention variability, we design scenarios spanning routine, goal-directed, and safety-critical driving. All scenarios are parameterized by weather, time-of-day, road type, traffic density, and pedestrian behavior. Parameters are randomized at episode start. We consider three scenario classes:

\begin{enumerate}
    \item \textbf{Unscripted navigation.} Participants drive freely in mixed urban-suburban environments. They choose their own routes and maneuvers.
    \item \textbf{Goal-directed navigation.} Participants follow audio-guided navigation through a series of checkpoints such as road construction, pedestrian crosswalks, multilane roundabouts, highway merges, etc.
    \item \textbf{Safety-critical events.} We inject five common near-miss situations with randomized timing, actor types, and traffic density:
    \begin{itemize}
        \item Highway emergency braking: sudden lead-vehicle deceleration with blocked adjacent lanes.
        \item Highway merging: entering dense traffic from the right lane.
        \item Urban pedestrian crossing: partially occluded pedestrian enters roadway.
        \item Signalized left turns with oncoming-vehicles.
        \item Highway cut-in with short-headway insertion by front vehicle. 
    \end{itemize}
\end{enumerate}

\begin{figure}
    \centering
    \includegraphics[width=0.9\linewidth]{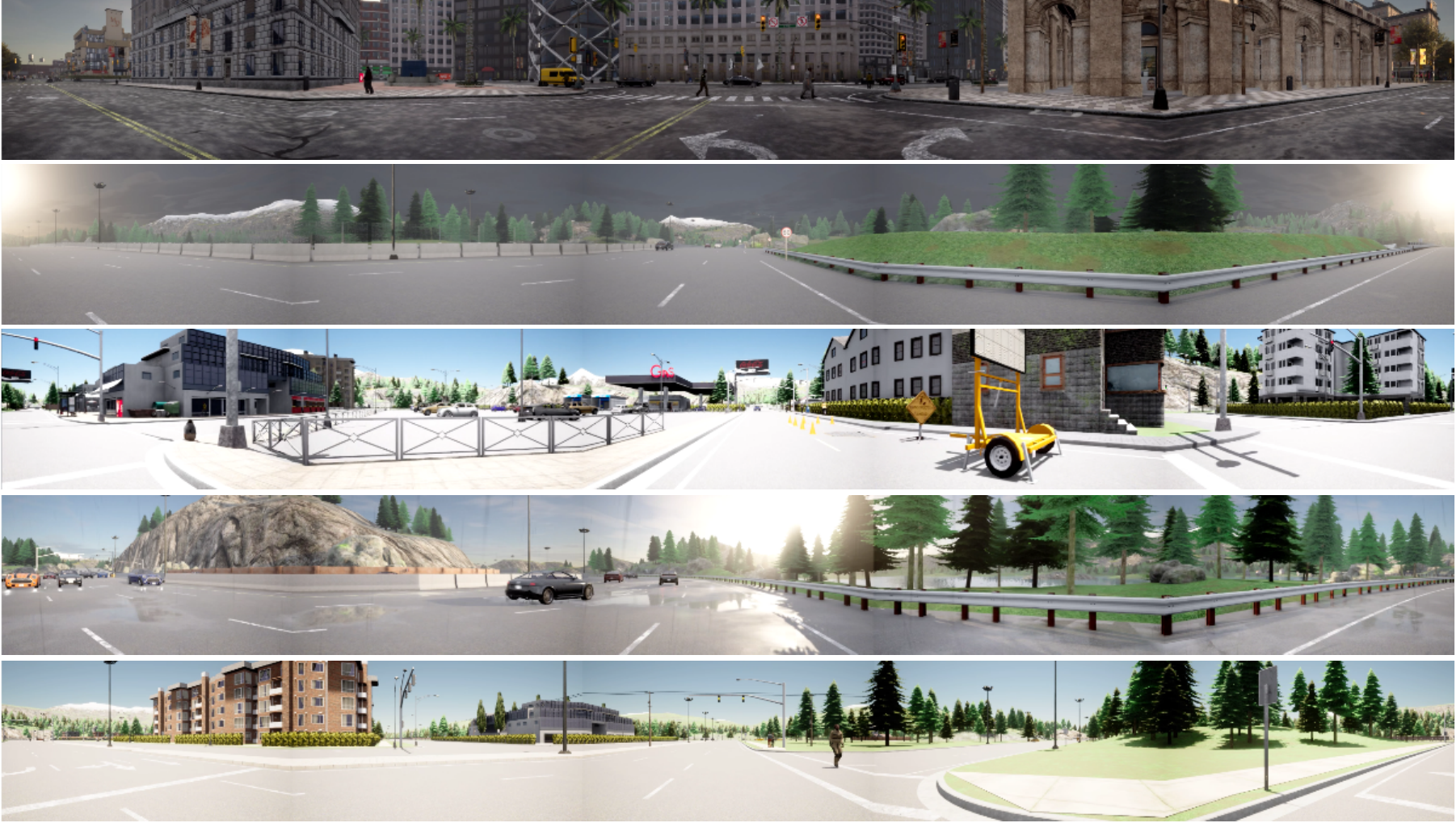}
    \caption{DriverGaze360 samples with different weather and traffic conditions.}
    \label{fig:dataset_samples}
\end{figure}

Across all categories, scene layouts, actor trajectories, and distractors are varied to increase behavioral coverage while maintaining reproducibility.~\Cref{fig:dataset_samples} shows data samples with varying weather and traffic conditions. More information regarding scenarios is provided in~\Cref{sec:traffic_scenarios} where~\Cref{fig:usecases} illustrates the safety-critical scenarios that were implemented.


\subsection{Fixation Target Calibration}

We align eye-tracker fixation points from the eye tracker coordinate frame to the CARLA simulator using AprilTags~\cite{olson2011tags}, shown in~\Cref{fig:experiment_running} (left). Each gaze point is mapped to the CARLA image plane via a homography. Similarly to prior-work~\cite{drveeye}, we build a fixation map by aggregating gaze within a 30-frame temporal window centered at time $t$: gaze points from neighboring frames are projected into frame $t$, converted to 2-D Gaussians with fixed spatial variance, and finally normalized to form a probability map,~\Cref{fig:experiment_running} (right). More details are provided in~\Cref{sec:fixation}.

\subsection{Participant Data Collection}

We recruited 21 licensed drivers (age 21--40) with at least three years of driving experience. Participants received a 5\,min on-boarding session to familiarize with the driving simulator and eye-tracking glasses. During data collection, participants were instructed to follow traffic regulations. Two subjects were excluded due to motion sickness or difficulty controlling the simulator.

Each session consisted of multiple driving episodes under varied conditions, including daylight, nighttime, wind, and rain, using standard CARLA weather presets. Simulator video streams were sampled at 30\,FPS to match the eye-tracker acquisition rate. 


\subsection{Dataset Properties}
\label{sec:properties}

\begin{figure}[thbp]
    \centering
    \includegraphics[width=\linewidth]{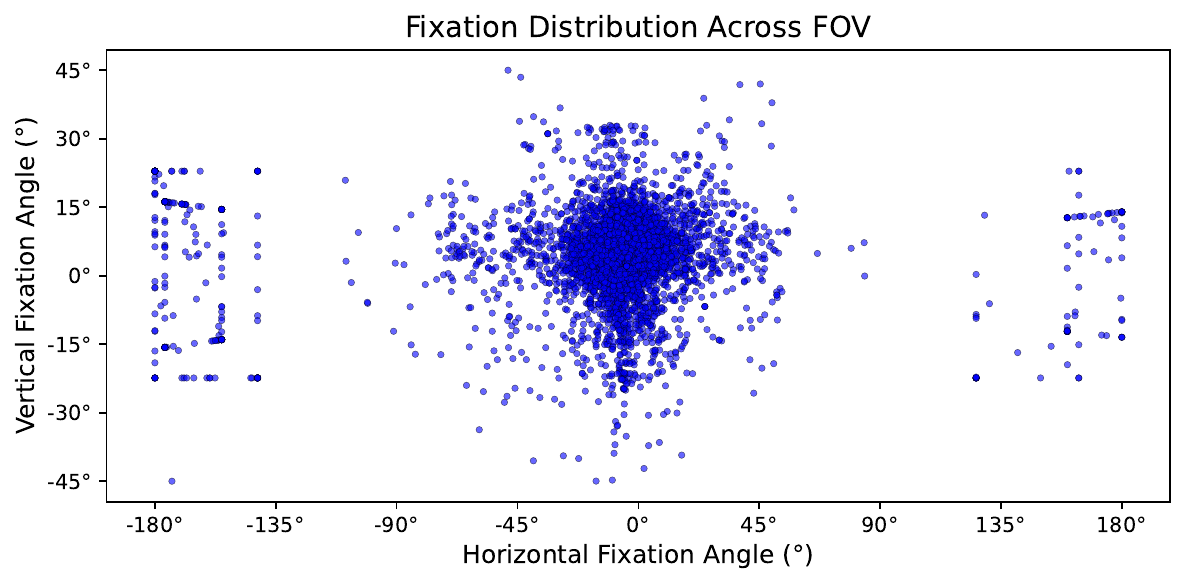}
    \caption{Fixation Distribution for DriverGaze360}
    \label{fig:gaze_dist}
\end{figure}

We introduce DriverGaze360, the first large-scale 360$^\circ$ driver attention dataset that provides dense gaze annotations, human driving behavior, and multi-sensor outputs. By leveraging CARLA's simulation and scenario replay capabilities, the dataset supports both traditional RGB-based gaze prediction and multi-sensor fusion approaches.\\
\textbf{Dataset statistics:} DriverGaze360 contains approximately 9\,h of driving footage with 1\,M gaze-annotated frames (obtained from 5 cameras with 1\,M frame each), collected from 19 participants, with per-driver contributions ranging from 18K to 126K frames. Data were recorded across three scenario types under diverse environmental conditions (day/night, rain, wind): $\sim$80 minutes of unscripted navigation, $\sim$85 minutes of safety-critical events, and $\sim$370 minutes of goal-directed navigation. Rear-view fixations account for 6\% of all gaze samples, aligning with reported real-world rates of 5-10\%~\cite{biondi2025distraction,viewglances}. A visualization of the gaze distribution is shown in~\Cref{fig:gaze_dist}. Our dataset is sampled at 30\,Hz and for each timestep it outputs five synchronized RGB frames at $1280 \times 720$ resolution.\\
\textbf{Comparison to existing datasets:} As evident in~\Cref{tab:dataset_compare}, DriverGaze360 provides a full 360$^\circ$ field of view in a simulated environment, enabling comprehensive modeling of driver attention under both routine and challenging scenarios. Even though our data is collected within a simulated environment it involves real human driving, unlike BDDA~\cite{bdda} or DADA-2000~\cite{dada2000}, which were collected by participants watching driving videos without direct engagement in driving. Moreover, we provide 9 hours of driving footage, more than any previous dataset, while covering both regular driving and critical situations.\\
\textbf{Data Splits.} Data were collected in standard CARLA maps referred to as Towns, namely Towns \{1, 2, 3, 4, 5, 6, 7, 10, and 11\}. We follow a map-based split strategy: Towns \{2, 3, 4, 7, 10, 11\} for training and Towns \{1, 5, 6\} for validation. This enforces non-overlapping geographic layouts while maintaining comparable distributions of urban, suburban, and highway scenarios. The resulting dataset is balanced across the splits with respect to duration.

\section{DriverGaze360-Net}

Omnidirectional scene capturing introduces new settings and challenges to driver attention which  we systematically explore in this section. We introduce our attention prediction network, DriverGaze-Net and an attended object mask generation algorithm.

\subsection{Network Architecture}

\begin{figure*}[thbp]
    \centering
    \includegraphics[width=\linewidth]{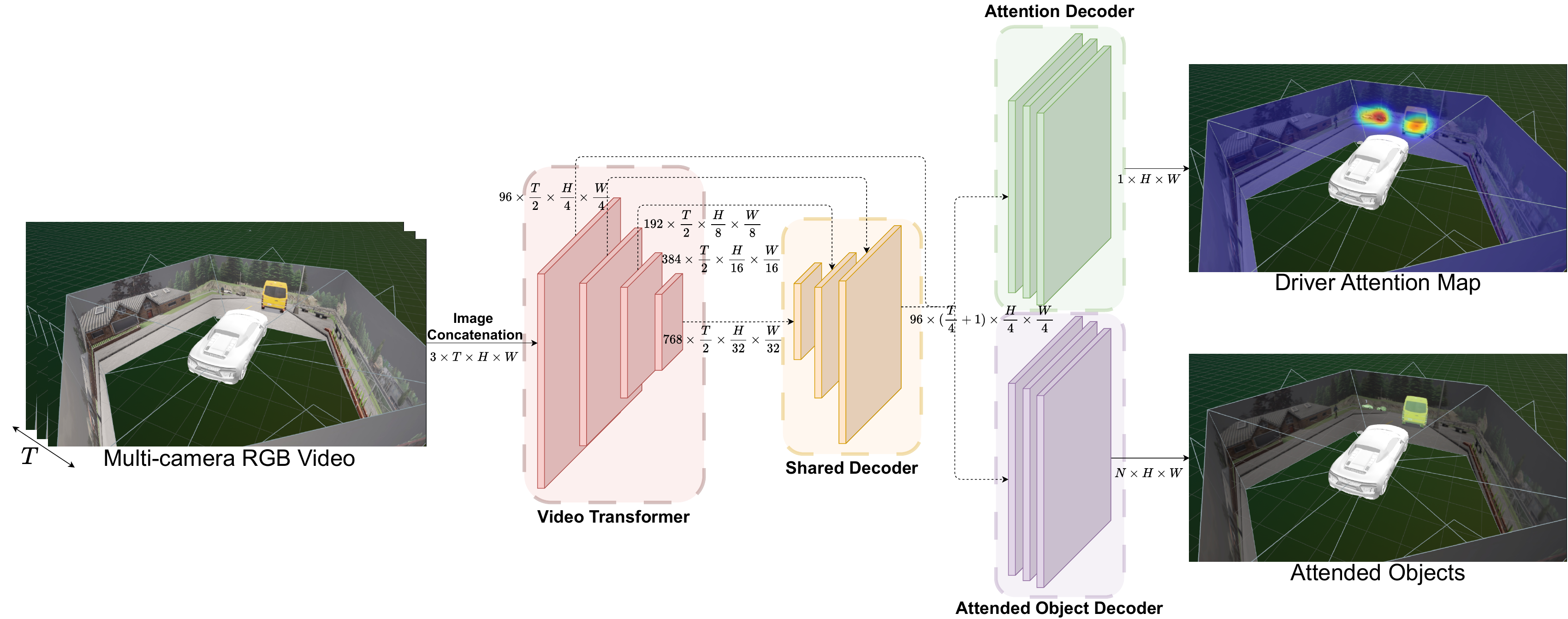}
    \caption{DriveGaze360-Net Architecture. We introduce a novel attended object prediction head that jointly learns an attention map and attended object segmentation. This addition improves the performance under wide panoramic images with highly sparse attention maps.}
    \label{fig:network_arch}
\end{figure*}

The proposed DriverGaze360-Net architecture, illustrated in Figure~\ref{fig:network_arch}, comprises three main components: a ViT-based~\cite{dosovitskiy2020image} scene encoder for spatio-temporal feature extraction, an attention decoder for driver attention prediction, and a segmentation decoder for attended objects. 

We adopt the Video Swin Transformer (VST)~\cite{videoswintransformer} as the backbone to extract multi-scale spatio-temporal features from sequences of RGB frames through the four encoder stages. VST was chosen because of its hierarchical architecture and shifted window-based self-attention mechanism~\cite{liu2021swin}. By performing attention calculations only within local, non-overlapping windows and shifting these windows between layers, it reduces computational complexity from quadratic (as with standard ViT) to linear with respect to image size.

The spatio-temporal features are fused through a shared convolutional decoder that progressively up-samples and aggregates encoder outputs using Conv3D layers and ReLU activations which enables weight sharing between downstream decoders. The attention decoder refines the shared representation using additional convolutional layers, consisting of Conv3D and ReLU activation blocks, with a final upsampling layer, to produce dense 360$^\circ$ gaze probability maps, while the attended object decoder mirrors this structure but includes a final classification layer to predict attended object categories and per-pixel scores, allowing object-level semantic reasoning aligned with driver attention. We define ``attended objects'' as active traffic participants (vehicles, pedestrians, cyclists), traffic signs and traffic lights that are perceived by the driver. 

The network operates on sequences of concatenated RGB frames from the past $T$ time steps, resulting in an input shape of $T \times 3 \times H \times W$. Features are first extracted via the VST encoder and then fed to the downstream decoders. The outputs of the network are an omnidirectional gaze attention map from the attention decoder and a segmentation map of attended objects from the attended object decoder.

\subsection{Attended Object Segmentation}

We jointly train a semantic segmentation head to predict attended objects, rather than treating attention prediction and semantic segmentation as independent tasks. Prior saliency networks such as~\cite{dadanet, drveeye} typically utilize full-scene semantic segmentation as input without explicitly distinguishing which objects are relevant to the driving context. This limitation becomes more pronounced in panoramic data, such as our DriverGaze360 dataset, where attention signals are inherently sparse and localized. Under such conditions, it is essential to focus the supervision on the attention-relevant objects rather than the entire scene.

To supervise object-aware attention, we generate new masks that retain only objects attended by the driver. Since most objects in 360$^\circ$ scenes are unseen at a given moment, full semantic supervision is not optimal, as we show in our ablation study~\Cref{sec:ablation_attented}. We first isolate road-user instances $R_{road} = $ \{vehicles, pedestrians, cyclists, traffic signs, traffic lights\}. The gaze map $S_{sal}$ is binarized to obtain a salient region mask, which is intersected with each instance. Instances with non-empty intersection are labeled as attended. We then construct an attended-object map $S_{obj}$ in which only pixels belonging to attended instances retain their semantic IDs; all others are set to background. The procedure is provided in Algorithm~\ref{alg:salient_selection}, where $\odot$ is element-wise multiplication and $\mathds{1}[\cdot]$ is an indicator function.

\begin{algorithm}[thbp]
\small{
\caption{Attended Object Extraction}
\label{alg:salient_selection}
\KwIn{Attention map $S_{sal}$, Instance segmentation $I_{inst}$, $\tau$ threshold, Road users $R_{road}$}
\KwOut{Semantic mask of attended objects $S_{obj}$}
    Extract road-user instances:
$I_{road} \leftarrow \{I_i \in I_{inst} \mid \text{class}(I_i) \in R_{road}\}$\;
    Binarize attention map: 
    $\hat{S}_{sal} \leftarrow \mathds{1}[S_{sal} > \tau]$\;
    Compute overlap:
    $M_{sal} \leftarrow \hat{S}_{sal} \odot I_{road}$\;
    Identify attended instances:
    $\mathcal{I}_{sal} \leftarrow \{I_i \in I_{road} \mid M_{sal} \cap I_i \neq \emptyset\}$\;
    Generate attended-object segmentation for all pixels~$p$:
$S_{obj}(p) = 
\begin{cases}
class\_id(I_i), & \text{if } p \in I_i, \ I_i \in \mathcal{I}_{sal}\\
0, & \text{otherwise}
\end{cases}$\;
\Return $S_{obj}$\;
}
\end{algorithm}

\subsection{Metrics}

Various metrics can be used to compare attention maps. We employ four widely-used saliency metrics~\cite{salmetrics}: Kullback-Leibler Divergence (KLD) that measure distance between two probability distributions, Correlation Coefficient (CC) which calculates the linear relationship between the ground truth and prediction, Similarity Index (SIM) which indicates the similarity, and Normalized Scanpath Saliency (NSS) which measures the difference, formulated in~\Cref{tab:metrics_saliency}. For semantic segmentation, we use two common metrics, Dice coefficient and Intersection-over-Union (IoU), as defined in~\Cref{tab:metrics_seg}.

\begin{table}[thbp]
    \small
  \centering
  \renewcommand{\arraystretch}{1.8}
  \begin{tabular}{c|c|c}
    \toprule    
    \textbf{Metric} & \textbf{Formulations} & \textbf{Range} \\
    \hline
    KLD $\downarrow$ & $\sum_i P_X(i) \log (\frac{P_X(i)}{P_Y(i) + \epsilon} + \epsilon)$ & $[0, \infty)$ \\
    \hline
    CC $\uparrow$ & $\frac{cov(P_{X}, P_{Y})}{\sigma(P_{X})\sigma(P_{Y})}$ & $[0,1]$ \\
    \hline
    NSS $\uparrow$ & $\frac{1}{N\neq0}\sum_i \bar{P}_Y(i) \cdot F(i)$ & $[0, \infty)$ \\
    \hline
    SIM $\uparrow$ & $\sum_i \min(P_X(i), P_Y(i))$ & $[0,1]$ \\
    \bottomrule
  \end{tabular}
  \caption{Metrics for attention map comparison. The symbols $\downarrow$ and $\uparrow$ indicate that lower or higher values are preferred, respectively. $P_{X}$ and $P_{Y}$ represent the ground-truth and predicted probability distributions of attention maps, respectively. $\bar{P}$ is the normalized probability distribution, $F$ denotes the fixation map for an image, $cov(P_X,P_Y)$ is the covariance, and $\sigma(\cdot)$ denotes the standard deviation. The summation index $i$ runs over image pixels, and $\epsilon$ is a small constant for numerical stability.}
  \label{tab:metrics_saliency}
\end{table}

\begin{table}[thbp]
    \small
  \centering
  \renewcommand{\arraystretch}{1.8}
  \begin{tabular}{c|c|c}
    \toprule    
    \textbf{Metric} & \textbf{Formulations} & \textbf{Range} \\
    \hline
    Dice $\uparrow$ & $2\frac{|X \cap Y|}{|X| + |Y|}$ & $[0, 1]$ \\
    \hline
    IoU $\uparrow$ & $\frac{|X \cap Y|}{|X \cup Y|}$ & $[0, 1]$ \\
    \bottomrule
  \end{tabular}
  \caption{Evaluation metrics for semantic segmentation. $X$ and $Y$ denote the sets of ground-truth and predicted pixels for a given class.} 
  \label{tab:metrics_seg}
\end{table}

\subsection{Loss Function}

Our model jointly optimizes attention prediction and attended object segmentation. For attention prediction, we use $\mathcal{L}_{sal}$ as defined in~\Cref{eq:salloss}, which combines the KL Divergence with the negative correlation coefficient between the ground-truth and predicted attention maps, $X_{sal}$ and $Y_{sal}$:
\begin{equation}
\small{
\mathcal{L}_{sal}(X_{sal}, Y_{sal}) = KLD(P_{X_{sal}}, P_{Y_{sal}}) - CC(X_{sal}, Y_{sal})
}
\label{eq:salloss}
\end{equation}

For attended object segmentation, the loss $\mathcal{L}_{seg}$ as defined in~\Cref{eq:segloss} maximizes the Dice and IoU scores while minimizing the cross-entropy \(\mathcal{L}_{CE} = -\sum_{c=1}^{N} y_c \log p_c\) between the ground-truth $X_{seg}$ and predicted $Y_{seg}$ segmentation maps:
\begin{equation}
\small{
\begin{split}
\mathcal{L}_{seg}(X_{seg},Y_{seg}) = - \ Dice(X_{seg}, Y_{seg}) - \ IoU(X_{seg},Y_{seg}) \\ + \ \mathcal{L}_{CE}(X_{seg}, Y_{seg})
\end{split}
}
\label{eq:segloss}
\end{equation}

The overall training objective is a weighted combination of these two losses, defined as $\mathcal{L}$ in~\Cref{eq:loss} with weights $\lambda_{sal}$ and $\lambda_{seg}$. In our experiments, we set $\lambda_{sal} = \lambda_{seg} = 1$:
\begin{equation}
\small{
    \mathcal{L}(X, Y) = \lambda_{{sal}} \mathcal{L}_{sal}(X_{sal}, Y_{sal}) + \lambda_{{seg}}\mathcal{L}_{seg}(X_{seg},Y_{seg})
    }
    \label{eq:loss}
\end{equation}

\section{Evaluation}
\label{sec:results}

In this section, we outline the experimental setup, present quantitative and qualitative results, and provide an in-depth analysis. We further assess real-world generalization using an external dataset and demonstrate the benefit of object-level guidance for attention prediction in an ablation experiment.

\subsection{Experimental Setup}

\textbf{Datasets.} For experiments on DriverGaze360, the input is formed by horizontally concatenating the five RGB camera views and resizing to $1120 \times 224$. Each training example consists of $16$ uniformly sampled consecutive frames ($\sim$0.5s), producing an input tensor of size $16 \times 3 \times 224 \times 1120$. Instance segmentation masks are extracted directly from CARLA. All evaluated methods are trained on the DriverGaze360 training set for fair comparison. Due to the absence of real panoramic driver attention datasets, we perform an experiment on the DADA-2000~\cite{dada2000} dataset with a narrow FoV to show that our method is widely applicable. For these evaluations, the images are resized to $224 \times 224$, and YOLO-v11~\cite{yolov11} is used for instance segmentation.\\
\textbf{Network.} We train our network, DriverGaze360-Net, using loss from ~\cref{eq:loss}, with the AdamW optimizer~\cite{loshchilov2017decoupled} (betas: $0.9$, $0.999$; weight decay: $0.01$) using a fixed learning rate of $1\times 10^{-4}$ and batch size of $4$. We use the Swin-S backbone pretrained on Kinetics-400~\cite{kay2017kinetics}. Our model is trained for 20 epochs with early stopping on a single NVIDIA H100 GPU, requiring approximately 24 hours. 


\subsection{Comparison with SOTA Methods}

\begin{figure}[thbp]
    \centering
    \includegraphics[width=0.85\linewidth]{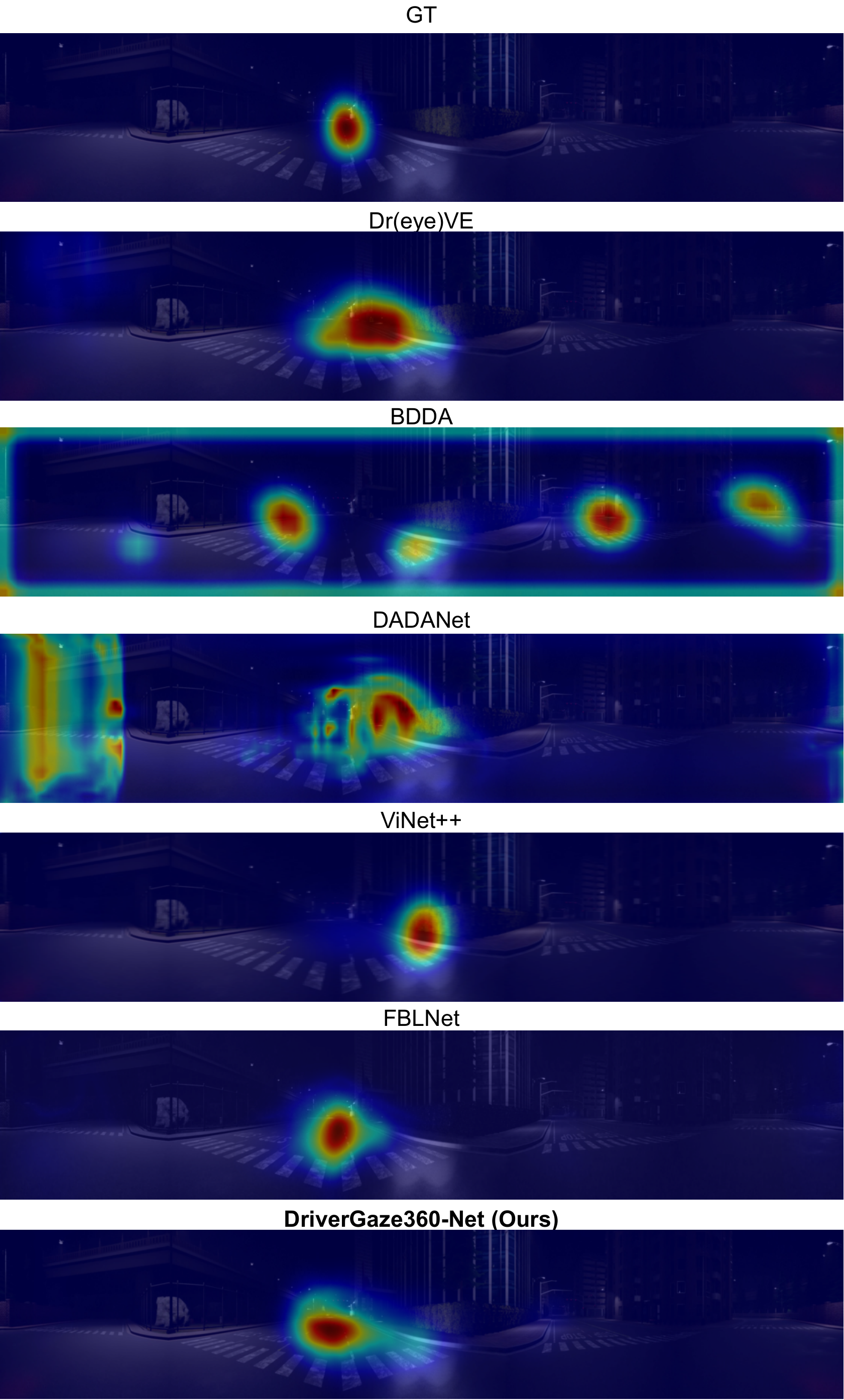}
    \caption{Qualitative results on DriverGaze360 (nighttime).}
    \label{fig:qualitative_drivergaze360_night}
\end{figure}

In order to illustrate the performance of our
method, we compared it with five other SOTA methods, among
which are Dr(eye)VE~\cite{drveeye}, BDDA~\cite{bdda}, DADANet~\cite{dadanet}, ViNet++~\cite{vinet2025}, and FBLNet~\cite{fblnet}. \\
\
\textbf{Quantitative Comparison.} We show the result on DriverGaze360 in~\Cref{tab:results_drivergaze360}. We
can observe that our method outperforms all other methods. Compared with the second-best result on KLD, SIM, CC and NSS metrics, our model achieves 12.18\%, 4.24\%, 4.51\%, 4.94\% performance improvement, respectively, demonstrating our method's advantage in wide panoramic inputs, which can also be seen qualitatively in~\Cref{fig:qualitative_drivergaze360_night}.


\begin{table}[thbp]
\small
  \centering
  \begin{tabular}{@{}lcccc@{}}
    \toprule
    \textbf{Model} & \textbf{KLD $\downarrow$} & \textbf{SIM $\uparrow$} & \textbf{CC $\uparrow$} & \textbf{NSS $\uparrow$} \\
    \midrule
    Dr(eye)VE~\cite{drveeye} & 1.293 & 0.340 & 0.613 & 5.452\\
    BDDA~\cite{bdda} & 2.566 & 0.218 &  0.400 & 2.772 \\
    DADANet~\cite{dadanet} & 1.269 & 0.476 & 0.618 & 5.598 \\
    ViNet++~\cite{vinet2025} & 1.251 & 0.442 & 0.611 & 2.772  \\
    FBLNet~\cite{fblnet} & \underline{1.215} & \underline{0.494} & \underline{0.639} & \underline{6.012} \\
    \hline
    DriverGaze360-Net (ours) & \textbf{1.067}	& \textbf{0.515} & \textbf{0.667}  & \textbf{6.309} \\
    \bottomrule
  \end{tabular}
  \caption{Results on DriverGaze360 Dataset.}
  \label{tab:results_drivergaze360}
\end{table}

\begin{figure}[thbp]
    \centering
    \includegraphics[width=.85\linewidth]{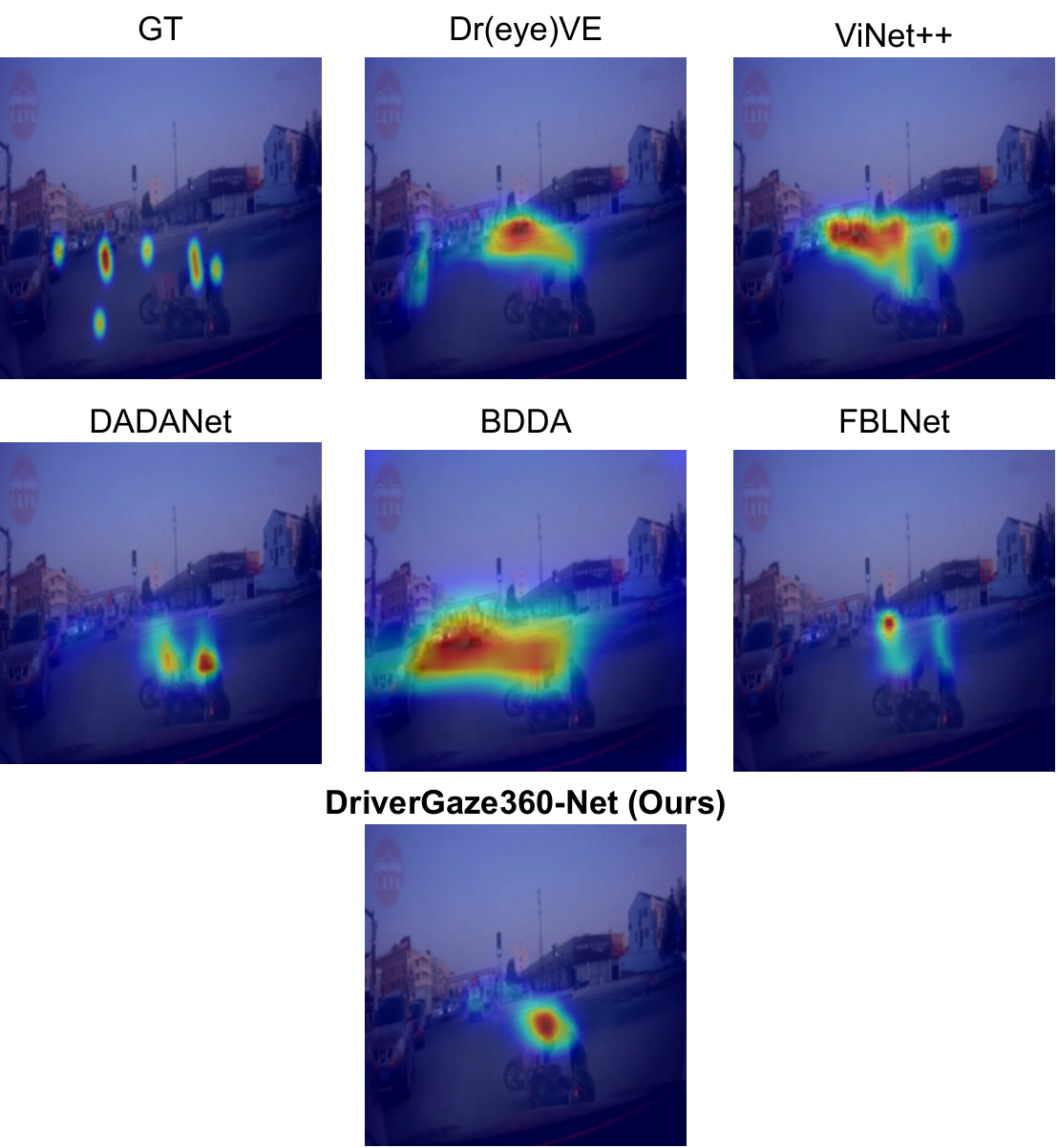}
    \caption{Qualitative results on DADA-2000 (motorbike accident).}
    \label{fig:qualitative_dada2000}
\end{figure}

\Cref{tab:results_dada} presents the results for the real-world DADA-2000~\cite{dada2000} dataset, here our method outperforms all baselines and improves on the second best method in SIM, CC, NSS by 7.32\%, 4.55\%, 3.36\% respectively, and is outperformed on KLD by just 0.48\%. This shows that our method is not dataset-specific and is applicable to real-world data as well, which is also demonstrated qualitatively in~\Cref{fig:qualitative_dada2000}.

\begin{table}[thbp]
    \small
  \centering
  \begin{tabular}{@{}lcccc@{}}
    \toprule
    \textbf{Model} & \textbf{KLD $\downarrow$} & \textbf{SIM $\uparrow$} & \textbf{CC $\uparrow$} & \textbf{NSS $\uparrow$}\\
    \midrule
    Dr(eye)VE~\cite{drveeye} & 2.065 & 0.325 & 0.451 & 2.920 \\
    BDDA~\cite{bdda} & 1.820 & 0.290 &  0.440 & 2.805  \\
    DADANet~\cite{dadanet} & \textbf{1.646} & 0.353 & \underline{0.484} & \underline{3.365} \\
    ViNet++~\cite{vinet2025} & 1.719 & 0.352 & 0.472  & 3.234 \\
    FBLNet~\cite{fblnet} & 1.818 & \underline{0.369} & 0.480 & 3.305 \\
    \hline
    DriverGaze360-Net (ours) & \underline{1.654} & \textbf{0.396} & \textbf{0.506} & \textbf{3.478} \\
    \bottomrule
  \end{tabular}
  \caption{Results on the DADA-2000 Dataset.}
  \label{tab:results_dada}
\end{table}

\begin{figure}[thbp]
    \centering
    \includegraphics[width=.87\linewidth]{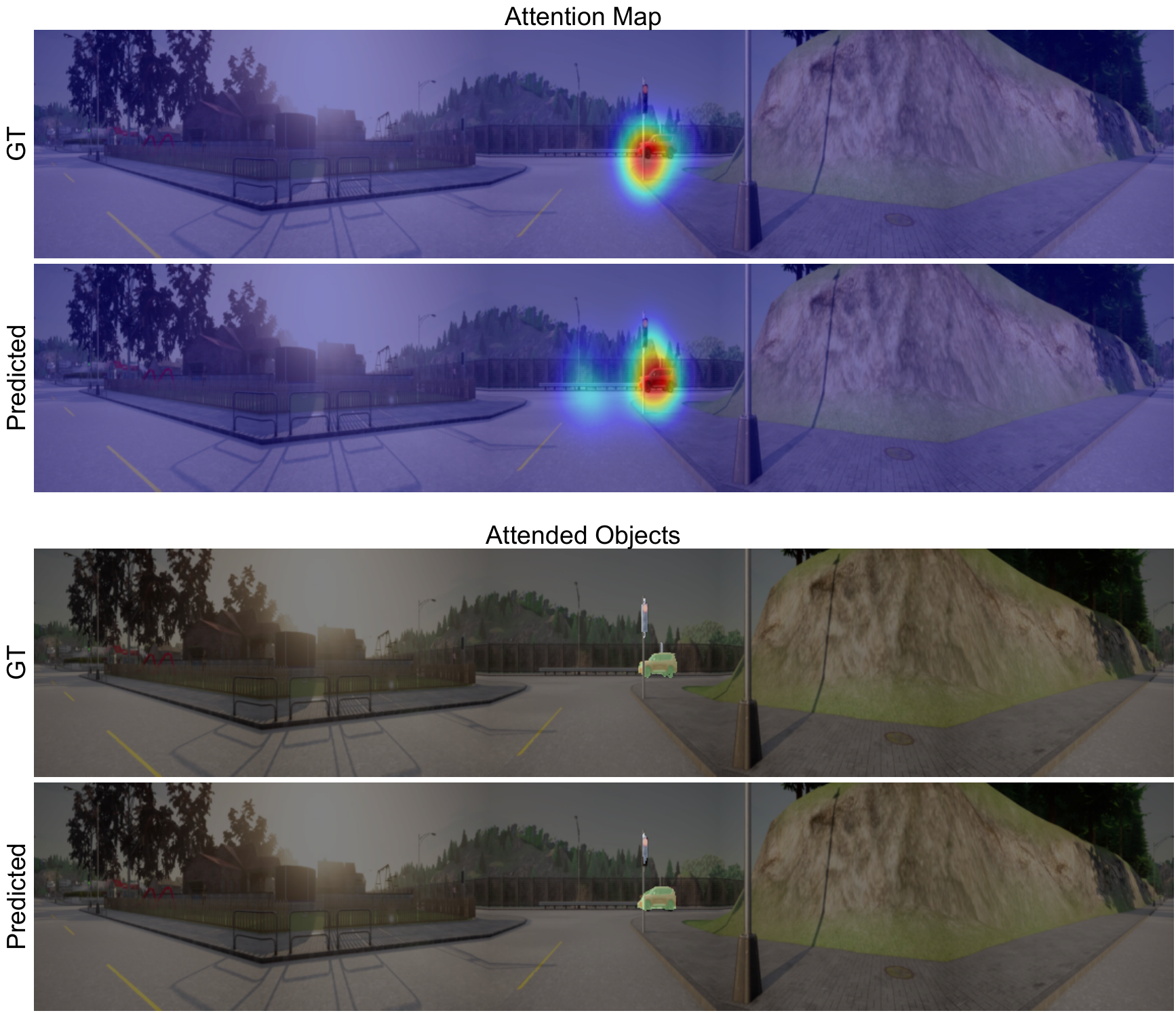}
    \caption{Attention prediction and Attended Object segmentation in DriverGaze360}
    \label{fig:attended_ours}
\end{figure}

\begin{figure}[thbp]
    \centering
    \includegraphics[width=0.7\linewidth]{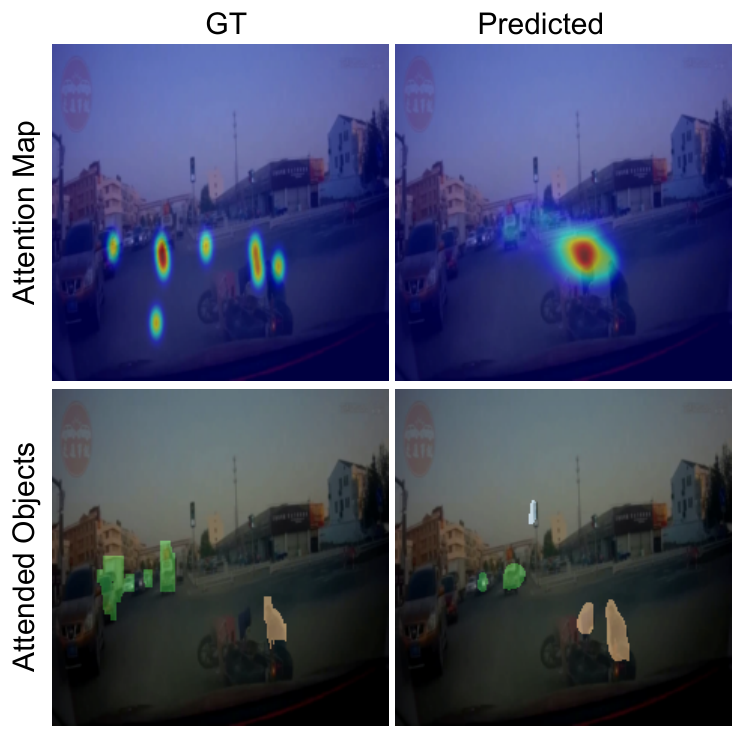}
    \caption{Attention prediction and Attended Object segmentation in DADA-2000 during a motorbike accident. }
    \label{fig:attended_dada}
\end{figure}

\subsection{Effect of Attended Object Segmentation}
\label{sec:ablation_attented}

\begin{table}[thbp]
\small
\centering
\setlength{\tabcolsep}{3.5pt}
\begin{tabular}{@{}lcccccc@{}}
\toprule
\textbf{Heads} & \textbf{KLD $\downarrow$} & \textbf{SIM $\uparrow$} & \textbf{CC $\uparrow$} & \textbf{NSS $\uparrow$} & \textbf{Dice $\uparrow$} & \textbf{IoU $\uparrow$} \\
\midrule
Attention  & 1.127 & 0.510 & 0.654 & 6.158  & - & - \\
+ ObjSeg. & \underline{1.092} & \underline{0.510} & \underline{0.659} & \underline{6.167} & \underline{0.636} & \underline{0.597} \\
+ AttObjSeg. & \textbf{1.067} & \textbf{0.515} & \textbf{0.667} & \textbf{6.309} & \textbf{0.639} & \textbf{0.626}  \\
\bottomrule
\end{tabular}
\caption{Ablation results on attended-object supervision.}
\label{tab:abalation_inputs}
\end{table}

\Cref{tab:abalation_inputs} reports the impact of adding the attended-object segmentation head. The baseline predicts only attention maps. Adding a semantic object segmentation head (ObjSeg), i.e. segmenting all objects regardless of the attention, yields consistent improvements, indicating that object-level cues help refine spatial attention. Introducing the attended-object segmentation head (AttObjSeg) further boosts all attention metrics, confirming that explicitly supervising the model to identify the viewer-relevant objects provides stronger guidance.

Relative to the attention-only baseline, AttObjSeg improves KLD, SIM, CC, and NSS by 5.32\%, 1.11\%, 2.08\%, and 2.45\%, respectively; compared to ObjSeg, it provides additional gains of 2.29\%, 1.08\%, 1.31\%, and 2.30\%. Qualitatively, the attended-object head successfully highlights task-critical regions (e.g., traffic lights, vehicles, and vulnerable road users), as shown in~\Cref{fig:attended_ours} and~\Cref{fig:attended_dada}.

\section{Conclusion}
\label{sec:conclusion}

In this work, we present a comprehensive study on omnidirectional driver attention, encompassing a new dataset, predictive algorithm, and extensive experimental validation. We introduced DriverGaze360, the first large-scale 360$^\circ$ driver attention dataset with dense gaze annotations, enabling detailed modeling of human visual behavior in full-surround driving environments. Building on this, we proposed a transformer-based model with an auxiliary semantic segmentation head that jointly predicts attention maps and attended objects, improving both spatial coherence and interpretability.
Our findings offer two key insights: 1) an omnidirectional dataset enables holistic modeling of driver awareness. 2) joint attended-object prediction significantly enhances panoramic attention estimation by guiding the model toward semantically meaningful regions.
For future work, we plan to leverage DriverGaze360 to develop explainable AI driver models that integrate human attention cues into autonomous decision-making, advancing transparency and reliability in autonomous vehicle systems~\cite{lmdrive, zhou2025towards, sima2024drivelm}.

\section*{Acknowledgments}

This work was partially funded by the European Union's Horizon Europe Research and Innovation Programme under Grant Agreement No. 101076360 (BERTHA) and by the German Federal Ministry of Research, Technology and Space under Grant Agreement No. 16IW24009 (COPPER).

The authors would like to express their sincere appreciation to Prateek Kumar Sharma, for his support with data collection and the implementation of driving scenarios. We also gratefully acknowledge Ruben Abad, Alex Levy, and Prof. Antonio M. López from the Computer Vision Center (CVC) for their methodological guidance and for providing the code used to implement the goal-directed navigation routes applied in collecting part of the dataset presented in this study. Finally, we sincerely thank all the participants who contributed to the dataset collection, as well as our colleagues at DFKI for their valuable feedback and support throughout this project. 
{
    \small
    \bibliographystyle{ieeenat_fullname}
    \bibliography{main}
}

\clearpage
\setcounter{page}{1}
\maketitlesupplementary

\section{Traffic Scenarios}
\label{sec:traffic_scenarios}


\subsection{Goal-Direction Navigation}

\begin{figure}[thbp]
    \centering
    \includegraphics[width=\linewidth]{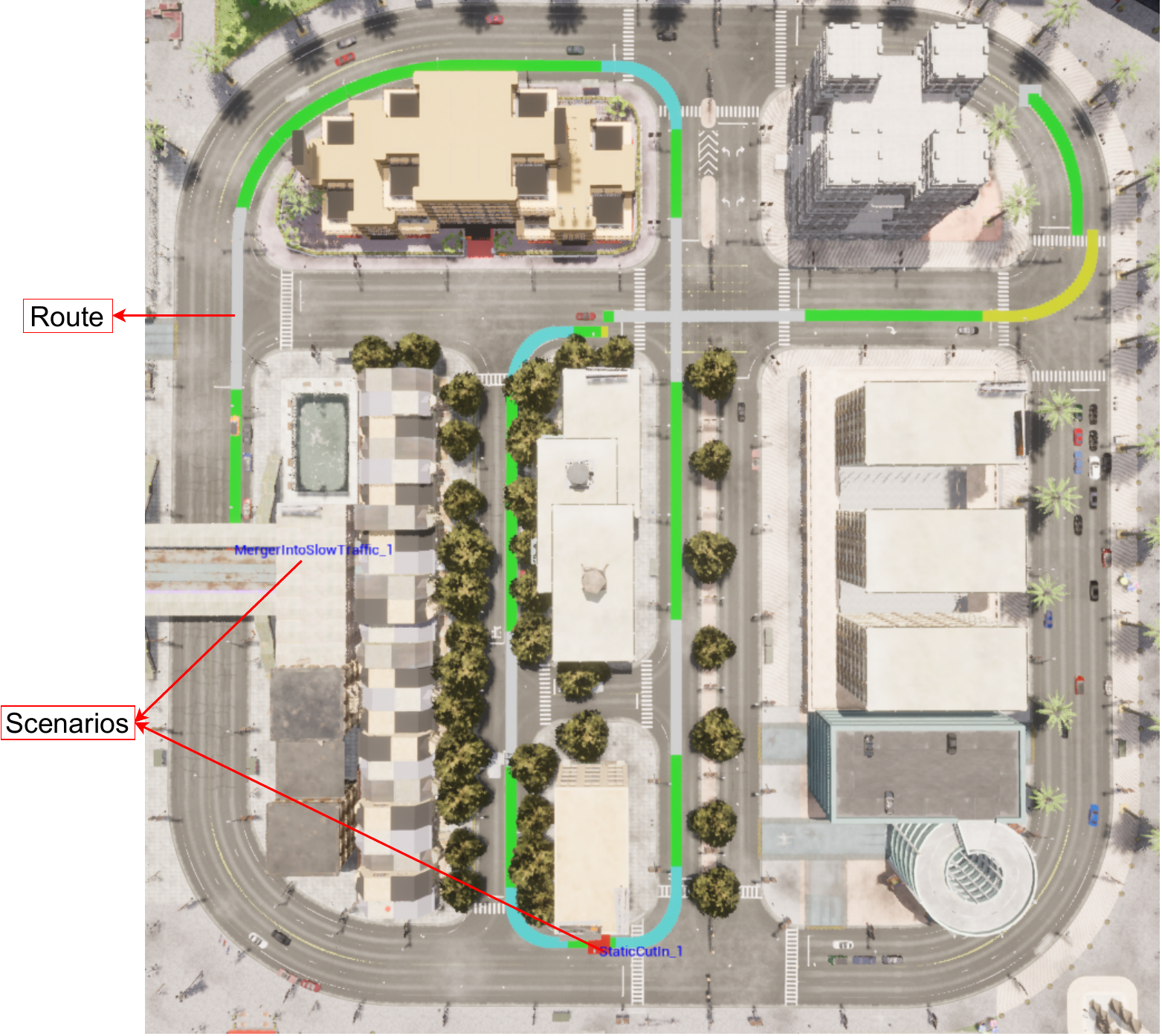}
    \caption{Example of Goal-Directed Navigation with two sub-scenarios.}
    \label{fig:goal_scenario}
\end{figure}

In goal-directed navigation driver participants follow a preplanned route using audio navigation cues (e.g., go straight, turn left, merge right), while interacting with regular city traffic and adhering to all traffic rules. Each session begins with randomized environmental conditions and progresses through diverse road types---including urban streets, highways, and multilane roundabouts. Session durations range from 7–15\,minutes, totaling to $\sim$370\,minutes for all sessions.~\Cref{fig:goal_scenario} shows one example of a goal-directed navigation session, consisting of 2 scripted sub-scenarios embedded within naturalistic driving. The range of scenarios varies from 2--10 depending on the route. 

\subsection{Safety-Critical Events}
\begin{figure}[thbp]
    \centering
    \includegraphics[width=\linewidth]{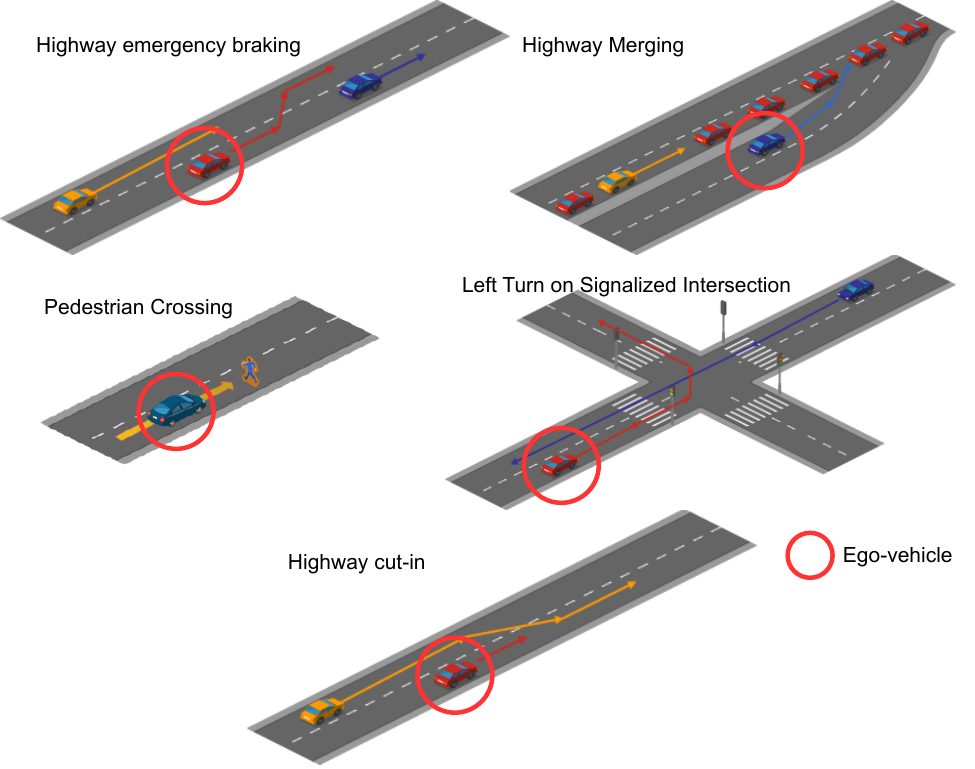}
    \caption{Safety-critical events.}
    \label{fig:usecases}
\end{figure}

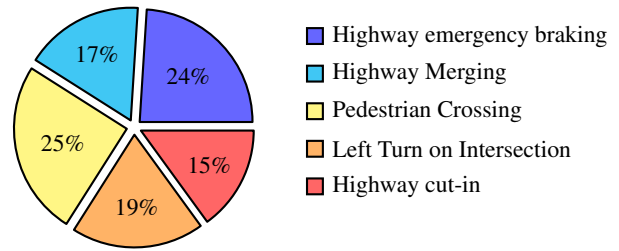
\begin{figure}[thbp]
\small
    \centering
    \begin{tikzpicture}
        \pie[radius=1.5, explode=0.1, text=legend]{24/Highway emergency braking,
        17/Highway Merging,
        25/Pedestrian Crossing,
        19/Left Turn on Intersection,
        15/Highway cut-in}
    \end{tikzpicture}
    \caption{Safety-critical event distribution. Total collected driving time in the dataset for safety-critical events is 85 minutes.}
    \label{fig:usecase_dist}
\end{figure}

\Cref{fig:usecases} summarizes the safety-critical scenarios integrated into the data collection pipeline. These events are implemented in SCENIC~\cite{fremont2023scenic} and parameterized by weather, map selection, traffic density, spawning locations, and actor configurations. Each event is executed as a short ($\leq$60s) self-contained clip. After every run, the simulator resets with a newly sampled parameter set, providing broad coverage across conditions. For each safety-critical scenario, we collect roughly 5–6 samples for each event from each driver. The distribution of collected examples across events is shown in~\Cref{fig:usecase_dist}.

We define these scenarios as:
\begin{itemize}
    \item \textbf{Highway emergency braking:} sudden deceleration from leading vehicle with blocked adjacent lanes in a highway.
    \item \textbf{Highway merging:} Highway merging with dense traffic from the right lane.
    \item \textbf{Urban pedestrian crossing:} Pedestrian crossing with partially occlusion in urban roadway.
    \item \textbf{Signalized left turn:} Turning left on a signalized intersection with oncoming-vehicles.
    \item \textbf{Highway cut-in:} Vehicle pull in front with a short-headway on a highway. 
\end{itemize}

\section{Eye-Tracker Fixation Alignment}
\label{sec:fixation}

\subsection{Fixation Calibration}

\begin{figure*}[t]
    \centering
    \includegraphics[width=\linewidth]{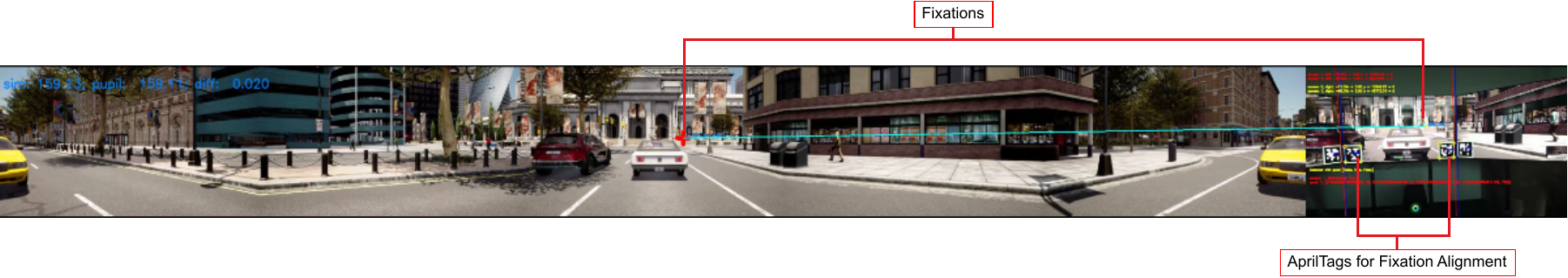}
    \caption{Fixation calibration between CARLA (left) and eye-tracker (right).}
    \label{fig:calibration}
\end{figure*}

\begin{figure*}[t]
    \centering
    \includegraphics[width=\linewidth]{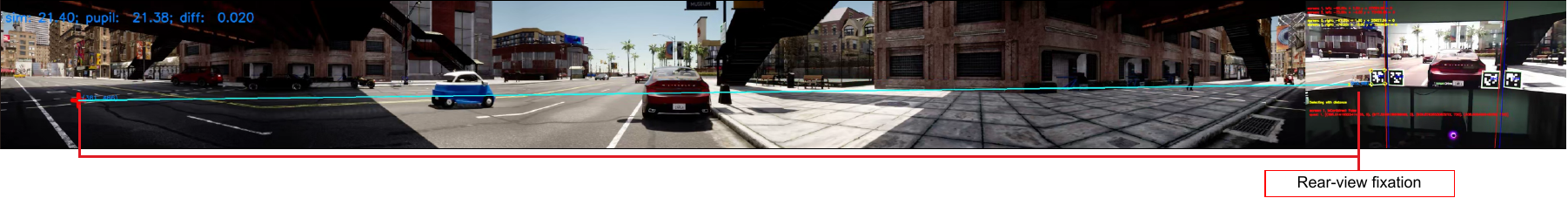}
    \caption{Fixation calibration for rear-view mirror. Driver checks the mirror before pulling out of parking.}
    \label{fig:calibration_rear}
\end{figure*}
We map fixation points from the eye-tracker frame to the CARLA coordinate frame using AprilTags~\cite{olson2011tags} and homography transformations. \Cref{fig:calibration} illustrates the calibration procedure for the forward view, while \Cref{fig:calibration_rear} shows the corresponding setup for the rear-view mirror.

We begin by detecting all AprilTags present in the image. Using the coordinates of each detected tag, we construct vertical lines from the left and right edges of the tag. These parallel lines allow us to determine which screen the user’s gaze currently falls on. Once the active screen is identified, we compute the homography between that screen and the CARLA simulator. The full procedure is detailed in \Cref{alg:fixation_calibration}.

\begin{algorithm}[thbp]
\small{
\caption{Fixation Calibration}
\label{alg:fixation_calibration}
\KwIn{Fixation coordinate in eye-tracker frame $X$, Egocentric image $I$, Simulator frame $F$}
\KwOut{Fixation coordinate in simulator frame $Y$}
    Extract AprilTags from $I$:
$T \leftarrow getAprilTags(I)$\;
    Get fixated screen $S$ using $X$ and $T$: 
    $S \leftarrow getCurrentScreen(X, T)$\;
    Calculate homography $H$ between $S$ and $F$:
    $H \leftarrow getHomography(S, F)$\;
    Compute Y:
    $Y \leftarrow H  X$\;
\Return $Y$\;
}
\end{algorithm}

\subsection{Attention Map Generation}

The driver attention map $S_{t}$ for a frame at time $t$ is built
by accumulating projected fixation points in a temporal sliding
window of $k = 30$ frames, centered at $t$. For each time step
$t + i$ in the window, where:
\[
i \in \left\{-\frac{k}{2}, -\frac{k}{2}+1, \ldots, \frac{k}{2}-1, \frac{k}{2}\right\},
\]
fixation point projections on $Y_{t+i}$ are estimated through the homography
transformation as discussed in~\Cref{alg:fixation_calibration}. A continuous fixation map is obtained from the projected fixations by centering
on each of them a multivariate Gaussian having a diagonal
covariance matrix $\sigma$:

\[
S_t(x, y) = 
\frac{1}{k}\sum_{i = -\frac{k}{2}}^{i = \frac{k}{2}}
\mathcal{N}\!\left( (x, y)\,\middle|\, Y_{t+i}, \sigma \right)
\tag{1}
\]

Eventually, each map $S_t$ is normalized to sum to 1, so that it forms a probability distribution of fixation points.

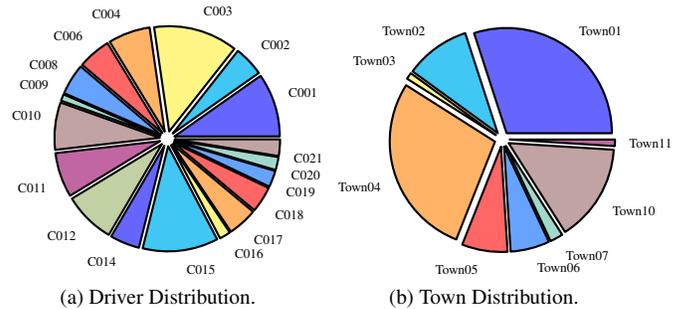
\begin{figure}[thbp]
\tiny
    \begin{subfigure}{0.48\linewidth}
    \centering
    \begin{tikzpicture}
        \pie[radius=1.4, explode=0.1, hide number]{
        9.76/C001,
        4.62/C002,
        12.91/C003,
        6.44/C004,
        5.05/C006,
        4.76/C008,
        1.02/C009,
        7.27/C010,
        6.85/C011,
        7.89/C012,
        4.54/C014,
        11.59/C015,
        1.70/C016,
        4.52/C017,
        4.00/C018,
        2.53/C019,
        2.00/C020,
        2.54/C021}
    \end{tikzpicture}
    \caption{Driver Distribution.}
    \label{fig:user_dist}
        
    \end{subfigure}%
    \hfill
    \begin{subfigure}{0.48\linewidth}
    \centering
    \begin{tikzpicture}
        \pie[radius=1.4, explode=0.1, hide number]{
        30/Town01,
        10/Town02,
        1/Town03,
        28/Town04,
        7/Town05,
        6/Town06,
        2/Town07,
        15/Town10,
        1/Town11}
    \end{tikzpicture}
    \caption{Town Distribution.}
    \label{fig:town_dist}
        
    \end{subfigure}
    \caption{DriverGaze360 Statistics.}
    \label{fig:stats}
    
\end{figure}

\begin{figure*}[t]
    \begin{subfigure}{\linewidth}
        \centering
        \includegraphics[width=\linewidth]{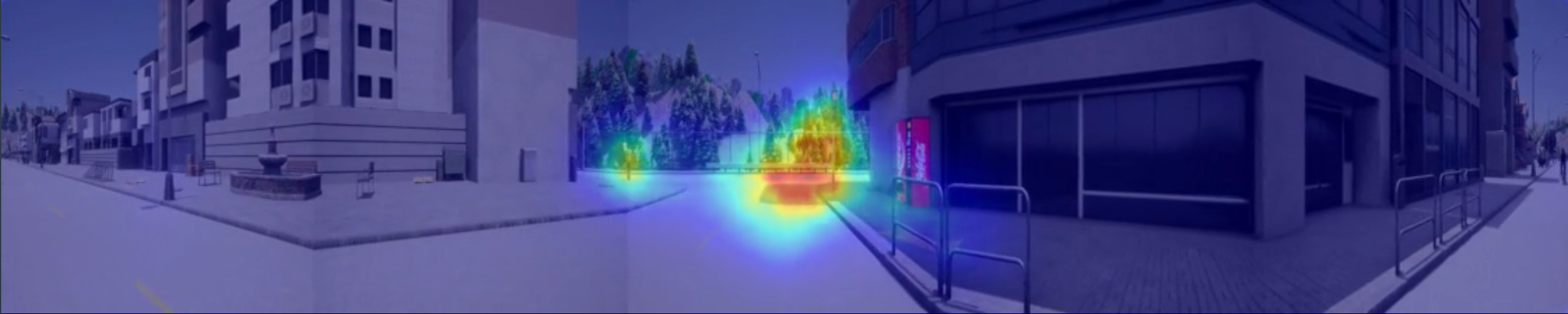}
        \caption{Simultaneous focus on traffic light and pedestrian.}
        \label{fig:supp_multiple}
    \end{subfigure}
    \begin{subfigure}{\linewidth}
        \centering
        \includegraphics[width=\linewidth]{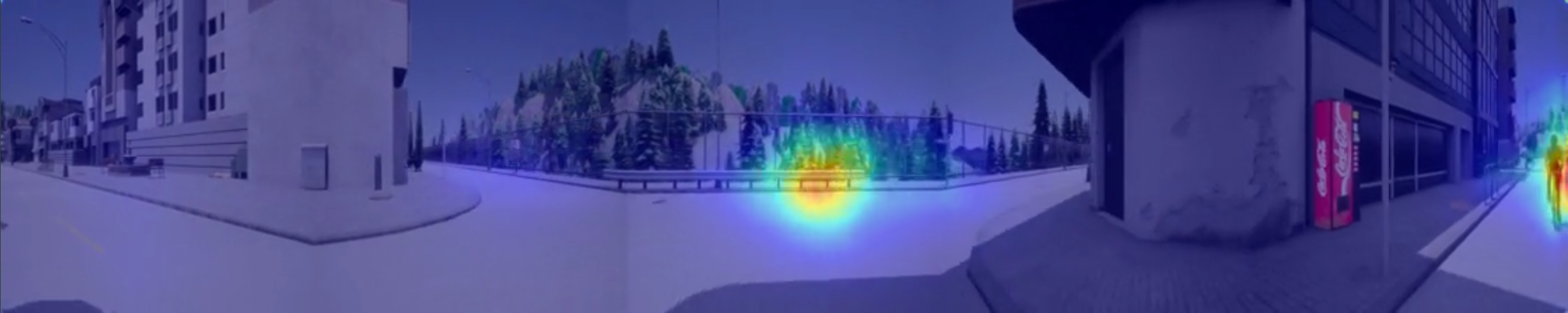}
        \caption{Focus on the cyclist in the rear-view mirror while turning.}
        \label{fig:supp_rear_cycle}
    \end{subfigure}
    \caption{DriverGaze360 Inference Results.}
\end{figure*}

\section{DriverGaze360 Statistics}

We present summary statistics from DriverGaze360 in~\Cref{fig:stats}. Driver contributions are fairly evenly distributed across the 19 participants (C001–C021, excluding C005 and C013). The contribution of each CARLA Town in the dataset is illustrated in~\Cref{fig:town_dist}. As described in~\Cref{sec:properties}, we partition the towns based on their geographic characteristics to ensure that no town appears in both the training and testing sets. After splitting, we balance the partitions so that the training set contains 303 minutes of footage (Towns 2, 3, 4, 7, 10, 11) and the validation set contains 234 minutes (Towns 1, 5, 6).

\section{Additional Qualitative Results}
Our method is simultaneously able to attend to the frontal view, traffic lights, and car in the rear-view. We demonstrate our method's ability to predict driver attention towards critical regions---such as pedestrians, cars, and traffic lights in~\Cref{fig:supp_multiple}. Moreover, it can predict attention to rear-view areas during turning maneuvers; for example, in~\Cref{fig:supp_rear_cycle}, the model correctly focuses on the cyclists while making a right-turn.

\section{Comparison to Panoramic Methods}

Uniquely to our setup, we use five rectilinear cameras (with no spherical distortion), not equirectangular projections typical of 360$^\circ$ saliency work. Nevertheless, we adapt two representative 360$^\circ$ saliency models~\cite{paver, martin20saliency} for our rectilinear input. As shown in~\Cref{tab:panoramic_comparison}, our method outperforms these adapted baselines.

\begin{table}[thbp]
    \small
  \caption{Comparison to Panoramic Methods.}
  \centering
  \begin{tabular}{@{}lcccc@{}}
    \toprule
    \textbf{Model} & \textbf{KLD $\downarrow$} & \textbf{CC $\uparrow$} & \textbf{SIM $\uparrow$} & \textbf{NSS $\uparrow$}\\
    \midrule
        PanoConv~\cite{martin20saliency} & \underline{1.450} & \underline{0.599} & \underline{0.425} & \underline{5.540} \\
        PAVER~\cite{paver} & 3.375 & 0.089 & 0.070 & 0.236 \\
    \hline
        DriverGaze360-Net (ours) & \textbf{1.067} & \textbf{0.667} & \textbf{0.515} &  \textbf{6.309} \\
    \bottomrule
  \end{tabular}
  \label{tab:panoramic_comparison}
\end{table}

\section{Limitations}

The primary limitation of DriverGaze360 is the sim-to-real gap inherent to simulation-based data collection. A simulator cannot perfectly replicate real-world driving conditions, which may influence participant behavior and gaze patterns relative to on-road driving. That said, simulation offers significant advantages: precise control over traffic and environmental conditions, and the ability to safely capture rare, high-risk events that are difficult to record in the real world. We therefore advise that claims about real-world generalization be interpreted cautiously, and encourage future work to investigate domain adaptation strategies to bridge this gap.

 \end{document}